\documentclass[a4paper]{article}

\usepackage{INTERSPEECH2015}

\usepackage{graphicx}
\usepackage{amssymb,amsmath,bm}
\usepackage{textcomp}

\ninept

\title{Long Short-Term Memory based Convolutional Recurrent Neural Networks for Large Vocabulary Speech Recognition}

\makeatletter
\def\name#1{\gdef\@name{#1\\}}
\makeatother \name{{\em Xiangang Li, Xihong Wu}}

\address{Speech and Hearing Research Center, \\
Key Laboratory of Machine Perception (Ministry of Education), \\
Peking University, Beijing, 100871\\
  {\small \tt \{lixg, wxh\}@cis.pku.edu.cn}
}

\begin{document}

  \maketitle
  \begin{abstract}
    Long short-term memory (LSTM) recurrent neural networks (RNNs) have been shown to give state-of-the-art performance on many speech recognition tasks, as they are able to provide the learned dynamically changing contextual window of all sequence history. On the other hand, the convolutional neural networks (CNNs) have brought significant improvements to deep feed-forward neural networks (FFNNs), as they are able to better reduce spectral variation in the input signal. In this paper, a network architecture called as convolutional recurrent neural network (CRNN) is proposed by combining the CNN and LSTM RNN. In the proposed CRNNs, each speech frame, without adjacent context frames, is organized as a number of local feature patches along the frequency axis, and then a LSTM network is performed on each feature patch along the time axis. We train and compare FFNNs, LSTM RNNs and the proposed LSTM CRNNs at various number of configurations. Experimental results show that the LSTM CRNNs can exceed state-of-the-art speech recognition performance.
  \end{abstract}

    \section{Introduction}

    Recently, the hybrid context dependent (CD) deep neural network (DNN) hidden Markov model (HMM) (CD-DNN-HMM) has become the dominant framework for acoustic modeling in speech recognition (e.g. \cite{AmDBN}\cite{CdDnn2}\cite{DnnForAm}). The performance improvement over the conventional Gaussian mixture model (GMM)-HMM is partially attributed to the powerful potential of DNN in modeling complex correlations in acoustic features.

    Based on the hybrid CD-DNN-HMM framework, many researches have been done from various aspects, such as the sequence discriminative training (e.g. \cite{Lattice1}\cite{Lattice2}\cite{Lattice3}), the network architectures (e.g. \cite{CNN.PhoneRecognition}\cite{LSTM.PhoneRecognition}\cite{LSTM.LargeScale}), and speaker adaptive methods (e.g. \cite{DNN.Adaptive.Ivector}\cite{DNN.Adaptive.SpeakerCode}), and have been shown to give significant performance improvements. In the researches of network architectures, two architectures have attracted lots of attentions: one is convolutional neural networks (CNNs), and the other is long short-term memory (LSTM) based recurrent neural networks (RNNs). In the seminal work, Ossama et al. \cite{CNN.PhoneRecognition} proposed to apply CNNs in the frequency domain to explicitly normalize speech spectral features to achieve frequency invariance and enforce locality of features, which have shown that further error rate reduction could be obtained comparing to the fully-connected DNNs on the phoneme recognition task. Subsequently, researchers have applied this idea on large vocabulary speech recognition tasks \cite{CNN.ASR.Trans}\cite{CNN.LVCSR}\cite{CNN.LVCSR.Pooling}. On the other hand, Graves et al.\cite{LSTM.PhoneRecognition} proposed to use stacked bidirectional LSTM network trained with connectionist temporal classification (CTC) \cite{CTC} for phoneme recognition. Subsequently, LSTM RNNs have been successfully applied and shown to give state-of-the-art performance on robust speech recognition task \cite{LSTM.RobustSpeechRecognition}, and many large vocabulary speech recognition tasks \cite{LSTM.LargeScale}\cite{LSTM.TIMIT.Bidirectional}\cite{LSTM.LVSR}\cite{LSTM.Deep}\cite{LSTM.SeqTraining}.

    In the literatures, most CNNs were applied on the frequency domain, and the variability along the time axis is handled by the fixed long time contextual window \cite{CNN.PhoneRecognition}\cite{CNN.ASR.Trans}\cite{CNN.LVCSR}. However, one of the original motivations for RNNs approach is to learn how much context to should be used for each prediction rather than fixed contextual window. Therefore, using recurrent connections in RNNs to improve CNNs is a natural choice. In this paper, an LSTM based convolutional recurrent neural network (CRNN) architecture is proposed by combining CNNs and LSTM RNNs. In the proposed approach, each speech frame, without adjacent context frames, is organized as a number of local feature patches along the frequency axis, and then a LSTM network is performed on each patch along the time axis. In other words, the proposed network architecture have convolutional operations to handle the variability along frequency axis, and recurrent operations to handle the variability along time axis. This proposed network architecture can be considered as introducing the recurrent operations in CNNs, or introducing the convolution operations in LSTM RNNs. Experiments are conducted on a large vocabulary conversational telephone speech recognition task, and results have shown that the proposed LSTM CRNNs can further improve the ASR performance. 


  \section{Review of LSTM RNNs and CNNs}

    In order to introduce the proposed network architecture, the conventional LSTM and CNN architectures for acoustic modeling are presented firstly in this section.

    \subsection{LSTM RNNs for acoustic modeling}
    In modern feed-forward neural networks (FFNNs) based hybrid acoustic modeling, acoustic context windows of 11 to 31 frames are typically used as inputs. The cyclic connections in RNNs exploit a self-learnt amount of temporal context, which makes them in principle better suited for acoustic modeling. The RNN-HMM hybrids have been studied for almost twenty years (e.g.\cite{RNN.Robinson1994}\cite{RNN.Povey2012}\cite{RNN.TIMIT}), and have been shown to give the state-of-the-art performance on many ASR tasks by introducing LSTM RNNs recently.

    Given an input speech sequence $x=(x_1, x_2, \ldots, x_T)$, an conventional RNN computes the hidden vector sequence $h=(h_1, h_2,\ldots, h_T)$ and output vector sequence $y=(y_1, y_2, \ldots, y_T)$ from $t=1$ to $T$ as
    \begin{align}\label{RNN}
    &h_t=\mathcal{H}(W_{xh}x_{t}+W_{hh}h_{t-1}+b_{h})\\
    &y_t=W_{hy}h_{t}+b_{y}
    \end{align}
    where, $W$ denotes weight matrices, $b$ denotes bias vectors and $\mathcal{H}$ denotes hidden layer function. However, RNNs are hard to be trained properly due to the vanishing gradient and exploding gradient problems as described in \cite{LRNN}. To address these problems, long short-term memory (LSTM) is proposed \cite{LSTM1}.

    \begin{figure}[htb]
      \centering
	  \includegraphics[width=65mm]{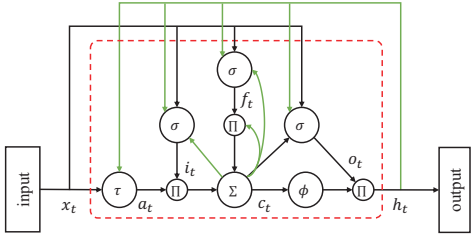}
      \caption{{\it The architecture of LSTM networks with one memory block, where green lines are time-delayed connections.}}
      \label{Lstm.Figure}
    \end{figure}

    The modern LSTM RNN architecture \cite{LSTM1}\cite{LSTM.Forget.Gate}\cite{LSTM.Peephole.Weights} is shown in Figure~\ref{Lstm.Figure}. In LSTM RNN, the recurrent hidden layer consists of a set of recurrently connected subnets known as ``memory blocks''. Each memory block contains one or more self-connected memory cells and three multiplicative gates to control the flow of information. Besides, there are peephole weights connecting gates to memory cell, which improve the LSTM¡¯s ability to learn precise timing and counting of internal states. The equations of LSTM memory blocks are as follows:
    \begin{align}\label{LSTM}
    &i_t=\sigma(W_{xi}x_t+W_{hi}h_{t-1}+W_{ci}c_{t-1}+b_{i})\\
    &f_t=\sigma(W_{xf}x_t+W_{hf}h_{t-1}+W_{cf}c_{t-1}+b_{f})\\
    &a_t=\tau(W_{xc}x_t+W_{hc}h_{t-1}+b_{c})\\
    &c_t=f_{t}c_{t-1}+i_{t}a_{t}\\
    &o_t=\sigma(W_{xo}x_t+W_{ho}h_{t-1}+W_{co}c_{t}+b_{o})\\
    &h_t=o_{t}\theta(c_t)
    \end{align}
    where, $\sigma$ is the logistic sigmoid function, and $i$, $f$, $o$, $a$ and $c$ are respectively the input gate, forget gate, output gate, cell input activation, and cell state vectors, and all of which are the same size as the hidden vector $h$. $W_{ci}$, $W_{cf}$, $W_{co}$ are diagonal weight matrices for peephole connections. $\tau$ and $\theta$ are the cell input and cell output non-linear activation functions, generally in this paper $tanh$. Besides, the LSTM Projected (LSTMP) network is proposed in \cite{LSTM.LargeScale}\cite{LSTM.LVSR}, which has a separate linear projection layer after the LSTM layer, and yield improved performance on a large vocabulary speech recognition task.

    \subsection{CNNs for acoustic modeling}
    CNN is capable of modeling local frequency structures by applying linear convolutional filters on the local feature patches representing a limited bandwidth of the whole speech spectrum. In order to represent speech inputs in a frequency scale that can be divided into a number of local bands, CNN for acoustic modeling always use the filter-bank features as the inputs. 
    Assuming the whole input feature is organized as $J$ local patches, and each patch $x_j(j=1,\ldots,J)$ has $s$ frequency bands, the equations of convolutional layer can be described as follow:
    \begin{equation}
    h_j = \theta(Wx_j + b), (j=1,\ldots,J)
    \label{eq.cnn}
    \end{equation}
    Where, $\theta(\cdot)$ is the activation function, $h_j$ is the convolutional layer¡¯s output vector of the $j$th feature patch. For each feature patch, the convolutional filter map the $s$ input nodes into $K$ output nodes, and the weights in convolutional filter are shared among all the feature patches.

    On top of each convolutional layer, a pooling layer is added to compute a lower resolution representation of the convolutional layer activations through sub-sampling. Usually, the max pooling function can be used as the pooling strategy, and in literature \cite{CNN.LVCSR.Pooling}, variants of pooling functions, such as the $l_p$ pooling \cite{CNN.Pooling.Lp}, stochastic pooling \cite{CNN.Pooling.Stochastic} were also evaluated.

    Both the weight sharing and pooling are important concepts in CNNs which helps to reduce the spectral variance in the input features. On top of stacked convolution-pooling layers, the standard fully connected layers are always added to combine the features of different bands.

    \section{LSTM based convolutional recurrent neural networks}
    From the descriptions of the LSTM RNNs and CNNs, we can find that, the LSTM RNNs provide the dynamically changing contextual window, while the weight sharing and pooling in CNNs focus on the frequency shift invariance. Motivated from taking both advantages, in this paper a new network is proposed which attempts to combine these properties from CNNs and LSTM RNNs.

\begin{figure}[htb]
    \centering
	\includegraphics[width=80mm]{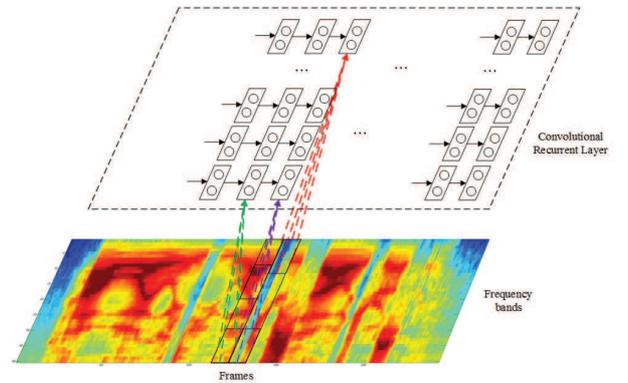}
    \caption{{\it Diagram showing the convolutional recurrent layer.}}
    \label{fig:CLstm.Figure}
    \end{figure}

    The convolutional layer in CNNs can be viewed as standard neural network layer operated on the local patches along the frequency axis re-organized from the input feature. In addition, a structure called ``Network In Network'' (NIN) is proposed in \cite{NIN} to enhance model for local patches within the receptive field, which replace the filters in conventional CNNs with a ``micro network'', such as a multilayer perceptron consisting of multiple fully connected layers with nonlinear activation functions. Based on these understanding of CNNs, and in order to combining CNNs and RNNs, the proposed network is constructed by replacing the filters in conventional CNNs with recurrent networks, specifically LSTM networks, which leads to the architecture illustrated in Figure~\ref{fig:CLstm.Figure}. This proposed network is called as convolutional LSTM (CLSTM) or LSTM based convolutional recurrent neural network (CRNN) in this paper.


    As illustrated in Figure~\ref{fig:CLstm.Figure}, the speech is represented with Mel-scale log-filterbank coefficients. Each speech frame, without context frames, is organized as local patches along the frequency axis, and adjacent patches have overlaps. Each patch represents a limited bandwidth of the whole speech spectrum, and a recurrent network is performed on each patch along the time axis. It means that, for each patch, a recurrent network receives the previous outputs and current inputs in the patch to make decisions. The equation of recurrent network based CRNNs can be written as:

    \begin{equation}
    h_{j,t}=\mathcal{H}(W_{xh}x_{j,t}+W_{hh}h_{j,t-1}+b_{h}), (j=1,\ldots,J)
    \label{eq.crnn}
    \end{equation}
    where, $h_{j,t}$ denotes the outputs of the $j$th patch in time $t$. For the LSTM CRNNs, we only need to change the hidden layer function, just like from conventional RNNs to LSTM RNNs.

    In the proposed CLSTM, like in the CNNs, the inputs of network are organized as a number of local feature patches. Meanwhile, same as in the LSTM RNNs, the input of the network only need current feature frame, without adjacent context frames. It is easy to find out that, there are convolution operations along the frequency axis, and recurrent operations along the time axis. The frequency shift invariance embodies in the convolutional part, while the dynamically changing contextual window embodies in the recurrent part.

    A similar network architecture to CLSTM is the multi-dimensional LSTM \cite{LSTM.MultiDimensions}. Through comparing these two architectures, it can be found out that, the CLSTM does not apply the recurrent operation along the adjacent frequency bands, while the multi-dimensional LSTM does. 
    Another related work is introduced in \cite{LSTM.CNNInput} on biological sequence data analyzing, where the network architecture is a 1-dimensional convolutional layer followed by an LSTM layer, a fully connected layer and a final softmax layer, which can be understood as the stack of convolutional layer and LSTM layer.

  \section{Experiments and discussion}
    We train and compare FFNNs, LSTM RNNs and the proposed LSTM CRNNs on a large vocabulary speech recognition task - the HKUST Mandarin Chinese conversational telephone speech recognition \cite{HKUST}. The corpus (LDC2005S15, LDC2005T32) is collected and transcribed by Hong Kong University of Science and Technology (HKUST), which contains 150-hour speech, and 873 calls in the training set and 24 calls in the development set, respectively. In our experiments, around 3-hour speech was randomly selected from the training set, used as the validate set for network training. The original development set in the corpus was used as ASR test set, which is not used in the training or the hyper-parameters determination processes.

    \subsection{Experimental setup}

    The speech in the corpus is represented with 25ms frames of Mel-scale log-filterbank coefficients (including the energy value), along with their first- and second-order temporal derivatives. The FFNNs use concatenated features, which are constructed by concatenating the current frame with 5 frames in its left and right contexts. The inputs to the LSTM RNNs and LSTM CRNNs are only the current frames (no window of frames).

    A trigram language model estimated using all the acoustic model training transcriptions is used in all the experiments. The hybrid approach \cite{CdDnn2}\cite{LSTMRNNAM} is used, in which the neural networks' outputs are converted as pseudo likelihood as the state output probability in hidden Markov model (HMM) framework. All the networks are trained based on the alignments generated by a well-trained GMM-HMM systems with 5529 senones (realignments by DNNs are not performed), and only the cross-entropy objective function is adopted.

    We implement the RNN training on multi-GPU devices. In the training, the truncated back-propagation though time (BPTT) learning algorithm \cite{BPTT} is adopted. Each sentence in the training set is split into subsequences with equal length (15 frames in the experiments), and two adjacent subsequences have overlapping frames (5 frames in the experiments). For computational efficiency, one GPU operates in parallel on 20 subsequences from different utterances at a time. In order to train these networks on multi-GPU devices, asynchronous stochastic gradient descent \cite{ASGD1}\cite{ASGD2} is adopted. The strategy introduced in \cite{RNN.difficulty} is applied to scale down the gradients. Since the information from the future frames helps making better decisions for current frame, we also delayed the output HMM state labels by 5 frames. In the experiments, the learning rate for training each network is decreased exponentially, and the initial and final learning rates are set specific to each network for stable convergence of training. 

    \subsection{Baseline systems}

    Firstly, the FFNNs and LSTM RNNs at various number of configurations are trained as the baseline, and results are summarized in Table 1 and Table 2. It is necessary to point out that, we found that, appropriate more senones would bring performance improvements. Thus, in this paper, we have 5529 senones against 3302 senones in \cite{LSTM.Deep}\cite{LSTM.Maxout}, leading to slightly better experimental results than that in \cite{LSTM.Deep}\cite{LSTM.Maxout}.

    In Table 1, ``4$\times$ReLU2000'' network has 4 hidden layers and each layer has 2000 rectified linear units (ReLU) \cite{ReLU}\cite{Relu2}, and ``4$\times$Maxout800G3'' network has 4 hidden layers and each layer has 800 maxout units \cite{MaxoutAm1}\cite{MaxoutAm2}, where the group size is 3. The CNN, denoted by ``2$\times$Conv+3$\times$ReLU2000'', has 2 convolution-pooling layers and 3 ReLU layers. In details, the convolutional layers has 256 units, and the pooling size is 3. It is expected that the CNN outperforms the other FFNNs.

    \begin{table} [t,h]
      \caption{\label{table1} {\it Character error rates (CER) of FFNNs based baseline systems on the HKUST speech recognition task.}}
      \vspace{2mm}
      \centerline{
      \begin{tabular}{|l|c|c|}
      \hline
      DNNs & \#Params & CER(\%) \\
      \hline \hline
      4$\times$ReLU2000 & 25.1M & 37.90 \\
      \hline
      4$\times$Maxout800G3 & 12.6M & 37.42 \\
      \hline
      2$\times$(Conv + Pooling) + 3$\times$ReLU2000 & 20.3M & \bf{36.66} \\
      \hline
      \end{tabular}}
    \end{table}

    \begin{table} [t,h]
      \caption{\label{table2} {\it Speech recognition results of LSTM RNNs based baseline systems on the HKUST speech recognition task.}}
      \vspace{2mm}
      \centerline{
      \begin{tabular}{|l|c|c|}
      \hline
      LSTM RNNs & \#Params & CER(\%) \\
      \hline \hline
      Lstm750 & 6.7M & 39.43 \\
      \hline
      Lstm2000P750 & 12.4M & 34.76 \\
      \hline
      Lstm750 + 3$\times$ReLU2000 & 23.1M & 36.46 \\
      \hline
      Lstm2000P750 + 3$\times$ReLU2000 & 28.8M & \bf{33.85} \\
      \hline
      3$\times$Lstm2000P750 & 39.4M & 34.22 \\
      \hline
      2$\times$Lstm2000P750 + 3$\times$ReLU2000 & 38.3M & 34.61 \\
      \hline
      2$\times$(Conv + Pooling) & & \\
      + 2$\times$ReLU2000 + Lstm2000P750 & 32.9M & 36.15 \\
      \hline
      Conv + Pooling + Lstm2000P750 & & \\
      + 3$\times$ReLU2000 & 44.4M & 36.01 \\
      \hline
      \end{tabular}}
    \end{table}

    In Table 2, ``Lstm750'' network has only 1 LSTM layer with 750 LSTM cells, and ``Lstm2000P750'' network has 1 LSTMP layer with 2000 LSTM cells projected to 750 nodes. Besides, based on the research in \cite{LSTM.Deep}, LSTM based deep RNNs are constructed.  LSTM RNNs yield better performance than FFNNs, and best performance among LSTM RNNs is obtained using ``Lstm2000P750+3$\times$ReLU2000'' network, which has an LSTMP layer followed by 3 ReLU layers. Besides, for comparison with the proposed LSTM CRNNs, we also trained networks construed by simply stacking convolutional layers and LSTM layers (the last two rows in Table~\ref{table2}), but unfortunately, these networks perform worse than ``Lstm2000P750+3$\times$ReLU2000''.

    \subsection{Results of CLSTMs}

    Since the pooling is a very importance concept in CNNs, we compared the models with and without pooling for the proposed CLSTMs, which shows no discernible performance difference. However, since the models with pooling layer have smaller number of parameters than that without pooling layer, the models in following experiments all have the pooling layers, and the pooling size is 3. Next, we explored the performance as a function of the number of LSTM cells for the convolutional recurrent layers. From Table 3, we can observe that as we increase the number of LSTM cells, the CER steadily decrease.  We were able to obtain a comparable performance by using 384 LSTM cells for the convolutional recurrent layer over the best baseline performance.

    \begin{table} [t,h]
      \caption{\label{table3} {\it Speech recognition results of the LSTM based convolutional recurrent networks.}}
      \vspace{2mm}
      \centerline{
      \begin{tabular}{|l|c|c|}
      \hline
      CLSTM RNNs & \#Params & CER(\%) \\
      \hline \hline
      CLstm128 + 3$\times$ReLU2000 & 25.3M & 36.00 \\
      \hline
      CLstm128 + Pooling + 3$\times$ReLU2000 & 21.2M & 35.92 \\
      \hline
      CLstm256 + Pooling + 3$\times$ReLU2000 & 23.4M & 34.81 \\
      \hline
      CLstm384 + Pooling + 3$\times$ReLU2000 & 25.3M & \bf{34.12} \\
      \hline
      \end{tabular}}
    \end{table}

    Literatures \cite{LSTM.LargeScale}\cite{LSTM.LVSR} have proposed the LSTMP to make more effective use of model parameters to train acoustic models. Similarly, we explored the projection layer strategy in CLSTM networks. Results of LSTMP CRNNs are shown in Table~4. The projection layer strategy seems to provide no performance improvements. However, by introducing projection layers, the total number of parameters with the same LSTM cells can be reduced. More specifically, with same LSTM cells, the CLSTMP network has similar performance with the CLSTM one, but smaller number of parameters, for example, the ``CLstm384+Pooling+3$\times$ReLU2000'' network and the ``CLstm384P128+Pooling+3$\times$ReLU2000'' network. When we increase the LSTM cells from 256 to 512 for the CLSTMP networks, there are only small changes in the total number of parameters, but obvious CER reductions.

    \begin{table} [t,h]
      \caption{\label{table4} {\it Speech recognition results of the LSTMP based convolutional recurrent networks.}}
      \vspace{2mm}
      \centerline{
      \begin{tabular}{|l|c|c|}
      \hline
      CLstmP RNNs & \#Params & CER(\%) \\
      \hline \hline
      CLstm256P128 + Pooling & & \\
      + 3$\times$ReLU2000 & 21.3M & 34.87 \\
      \hline
      CLstm384P256 + Pooling & & \\
      + 3$\times$ReLU2000 & 23.7M & 34.11 \\
      \hline
      CLstm512P256 + Pooling & & \\
      + 3$\times$ReLU2000 & 23.8M & \bf{34.03} \\
      \hline
      \end{tabular}}
    \end{table}

    In the literatures, many studies have shown that performance can be improved by using multiple LSTM layers. Besides, multiple convolutional layers can also improve CNNs \cite{CNN.LVCSR}\cite{CNN.LVCSR.Pooling}. Thus, experiments were conducted to explore having two convolutional recurrent layers or another recurrent layers in the LSTM CRNNs. Results with different network structure configurations are shown in Table~\ref{table5}. The table shows that having two convolutional recurrent layers also helps and yields a 3.8\% relative improvement performance over the baseline systems. What is noteworthy is that the networks that have another LSTMP layers on the top of CLSTM layer can further reduce the CER to 31.43\%, which is a 7.1\% relative improvement.

    \begin{table} [t,h]
      \caption{\label{table5} {\it Speech recognition results of networks with multiple convolutional recurrent layers or multiple recurrent layers.}}
      \vspace{2mm}
      \centerline{
      \begin{tabular}{|l|c|c|}
      \hline
      DNNs & \#Params & CER(\%) \\
      \hline \hline
      2$\times$(CLstm256 + Pooling) & & \\
      + 3$\times$ReLU2000 & 21.2M & 32.89 \\
      \hline
      2$\times$(CLstm384P256 + Pooling) & & \\
      + 3$\times$ReLU2000 & 22.3M & 32.54 \\
      \hline
      CLstm256P128 + Pooling & & \\
      + Lstm2000P750 + 3$\times$ReLU2000 & 36.4M & 32.34 \\
      \hline
      CLstm384P256 + Pooling & & \\
      + Lstm2000P750 + 3$\times$ReLU2000 & 44.9M & \bf{31.43} \\
      \hline
      \end{tabular}}
    \end{table}

    In summary, the experimental results show that the proposed CLSTM network can exceed state-of-the-art ASR performance. The best performance can be obtained by the network which is constructed by one CLSTMP layer, one LSTMP layer and three ReLU layers.

  \section{Conclusions}



    In this paper, an LSTM based convolutional recurrent neural network (CRNN) architecture is proposed for acoustic modeling by combining the CNNs and LSTM RNNs, which is constructed by replacing the filter in conventional CNNs with a recurrent filter, specifically a LSTM based filter. The proposed network can be considered as introducing the dynamically changing contextual window embedded in the LSTM network to the conventional CNNs, or introducing the frequency shift invariance embedded in the convolutional structure to LSTM RNNs. In other words, the proposed network contains convolutional operations along the frequency axis, and recurrent operations along the time axis.


    We empirically evaluated the proposed network against FFNNs and LSTM networks on a large vocabulary speech recognition task. In the experiments, various configurations for constructing deep networks have been compared. The experimental results revealed that, the proposed LSTM CRNN can further improve the performance, delivering a 7\% CER relative reduction significantly comparing to LSTM networks which have been shown to give state-of-the-art performance on some ASR tasks. However, we believe that this work is just a preliminary study. Future work includes training the CLTSM CRNNs using sequence discriminative training criterion \cite{LSTM.SeqTraining} and experiments on a larger corpus.



  \newpage
  \eightpt
  \bibliographystyle{IEEEtran}

  \bibliography{mybib}

\begin{thebibliography}{10}
\providecommand{\url}[1]{#1}
\csname url@samestyle\endcsname
\providecommand{\newblock}{\relax}
\providecommand{\bibinfo}[2]{#2}
\providecommand{\BIBentrySTDinterwordspacing}{\spaceskip=0pt\relax}
\providecommand{\BIBentryALTinterwordstretchfactor}{4}
\providecommand{\BIBentryALTinterwordspacing}{\spaceskip=\fontdimen2\font plus
\BIBentryALTinterwordstretchfactor\fontdimen3\font minus
  \fontdimen4\font\relax}
\providecommand{\BIBforeignlanguage}[2]{{%
\expandafter\ifx\csname l@#1\endcsname\relax
\typeout{** WARNING: IEEEtran.bst: No hyphenation pattern has been}%
\typeout{** loaded for the language `#1'. Using the pattern for}%
\typeout{** the default language instead.}%
\else
\language=\csname l@#1\endcsname
\fi
#2}}
\providecommand{\BIBdecl}{\relax}
\BIBdecl

\bibitem{AmDBN}
A.~Mohamed, G.~Dahl, and G.~Hinton, ``Acoustic modeling using deep belief
  networks,'' \emph{IEEE Trans. Audio Speech Lang. Processing}, vol.~20, pp.
  14--22, 2012.

\bibitem{CdDnn2}
G.~Dahl, D.~Yu, L.~Deng, and A.~Acero, ``Context-dependent pretrained deep
  neural networks for large-vocabulary speech recognition,'' \emph{IEEE Trans.
  Audio Speech Lang. Processing}, vol.~20, pp. 30--42, 2012.

\bibitem{DnnForAm}
G.~Hinton, L.~Deng, D.~Yu, G.~Dahl, A.~Mohamed, N.~Jaitly, A.~Senior,
  V.~Vanhoucke, P.~Nguyen, T.~Sainath, and B.~Kingsbury, ``Deep neural networks
  for acoustic modeling in speech recognition: the shared views of four
  research groups,'' \emph{IEEE Signal Processing Mag.}, vol.~29, pp. 82--97,
  2012.

\bibitem{Lattice1}
B.~Kingsbury, ``Lattice-based optimization of sequence classification criteria
  for neural-network acoustic modeling,'' in \emph{ICASSP}, 2009, pp.
  3761--3764.

\bibitem{Lattice2}
B.~Kingsbury, T.~Sainath, and H.~Soltau, ``Scalable minimum bayes risk training
  of deep neural network acoustic models using distributed hessian-free
  optimization,'' in \emph{Interspeech}, 2012, pp. 10--13.

\bibitem{Lattice3}
K.~Vesel{\'{y}}, A.~Ghoshal, L.~Burget, and D.~Povey, ``Sequence-discriminative
  training of deep neural networks,'' in \emph{Interspeech}, 2013, pp.
  2345--2349.

\bibitem{CNN.PhoneRecognition}
O.~Abdel-Hamid, A.~Mohamed, H.~Jiang, and G.~Penn, ``Applying convolutional
  neural networks concepts to hybrid nn-hmm model for speech recognition,'' in
  \emph{ICASSP}, 2012, pp. 4277--4280.

\bibitem{LSTM.PhoneRecognition}
A.~Graves, A.~Mohamed, and G.~Hinton, ``Speech recognition with deep recurrent
  neural networks,'' in \emph{ICASSP}, 2013, pp. 6645--6649.

\bibitem{LSTM.LargeScale}
H.~Sak, A.~Senior, and F.~Beaufays, ``Long short-term memory recurrent neural
  network architectures for large scale acoustic modeling,'' in
  \emph{Interspeech}, 2014, pp. 338--342.

\bibitem{DNN.Adaptive.Ivector}
G.~Saon, H.~Soltau, D.~Nahamoo, and M.~Picheny, ``Speaker adaptation of neural
  network acoustic models using i-vectors,'' in \emph{ASRU}, 2013, pp. 55--59.

\bibitem{DNN.Adaptive.SpeakerCode}
O.~Abdel-Hamid and H.~Jiang, ``Fast speaker adaptation of hybrid nn/hmm model
  for speech recognition based on discriminative learning of speaker code,'' in
  \emph{ICASSP}, 2013, pp. 7942--7946.

\bibitem{CNN.ASR.Trans}
O.~Abdel-Hamid, A.~Mohamed, H.~Jiang, L.~Deng, G.~Penn, and D.~Yu,
  ``Convolutional neural networks for speech recognition,'' \emph{IEEE/ACM
  Trans. Audio Speech Lang. Processing}, vol.~22, pp. 1533--1545, 2014.

\bibitem{CNN.LVCSR}
T.~Sainath, A.~Mohamed, B.~Kingsbury, and B.~Ramabhadran, ``Deep convolutional
  neural networks for lvcsr,'' in \emph{ICASSP}, 2013.

\bibitem{CNN.LVCSR.Pooling}
T.~Sainath, B.~Kingsbury, A.~Mohamed, G.~Dahl, G.~Saon, H.~Soltau, T.~Beran,
  A.~Aravkin, and B.~Ramabhadran, ``Improvements to deep convolutional neural
  networks for lvcsr,'' 2013, arXiv:1309.1501.

\bibitem{CTC}
A.~Graves, S.~Fern{\'{a}}ndez, F.~Gomez, and J.~Schmidhuber, ``Connectionist
  temporal classification: Labelling unsegmented sequence data with recurrent
  neural network,'' in \emph{ICML}, 2006, pp. 369--376.

\bibitem{LSTM.RobustSpeechRecognition}
J.~Geiger, Z.~Zhang, F.~Weninger, B.~Schuller, and G.~Rigoll, ``Robust speech
  recognition using long short-term memory recurrent neural networks for hybrid
  acoustic modelling,'' in \emph{Interspeech}, 2014, pp. 631--635.

\bibitem{LSTM.TIMIT.Bidirectional}
A.~Graves, N.~Jaitly, and A.~Mohamed, ``Hybrid speech recognition with deep
  bidirectional lstm,'' in \emph{ASRU}, 2013, pp. 273--278.

\bibitem{LSTM.LVSR}
H.~Sak, A.~Senior, and F.~Beaufays, ``Long short-term memory based recurrent
  neural network architectures for large vocabulary speech recognition,'' 2014,
  arXiv:1402.1128.

\bibitem{LSTM.Deep}
X.~Li and X.~Wu, ``Constructing long short-term memory based deep recurrent
  neural network for large vocabulary speech recognition,'' in \emph{ICASSP},
  2015.

\bibitem{LSTM.SeqTraining}
H.~Sak, O.~Vinyals, G.~Heigold, A.~Senior, E.~McDermott, R.~Monga, and M.~Mao,
  ``Sequence discriminative distributed training of long short-term memory
  recurrent neural networks,'' in \emph{Interspeech}, 2014, pp. 1209--1213.

\bibitem{RNN.Robinson1994}
A.~Robinson, ``An application of recurrent nets to phoneme probability
  estimation,'' \emph{IEEE Trans. Neural Networks}, vol.~5, pp. 298--305, 1994.

\bibitem{RNN.Povey2012}
O.~Vinyals, S.~Ravuri, and D.~Povey, ``Revisiting recurrent neural networks for
  robust asr,'' in \emph{ICASSP}, 2012, pp. 4085--4088.

\bibitem{RNN.TIMIT}
L.~Deng and J.~Chen, ``Sequence classification using the high-level features
  extracted from deep neural networks,'' in \emph{ICASSP}, 2014, pp.
  6844--6848.

\bibitem{LRNN}
Y.~Bengio, P.~Simard, and P.~Frasconi, ``Learning long-term dependencies with
  gradient descent is difficult,'' \emph{IEEE Trans. Neural Networks}, vol.~5,
  pp. 157--166, 1994.

\bibitem{LSTM1}
S.~Hochreiter and J.~Schimidhuber, ``Long short-term memory,'' \emph{Neural
  Computation}, vol.~9, pp. 1735--1780, 1997.

\bibitem{LSTM.Forget.Gate}
F.~Gers, J.~Schmidhuber, and F.~Cummins, ``Learning to forget: Continual
  prediction with lstm,'' \emph{Neural Computation}, vol.~12, pp. 2451--2471,
  2000.

\bibitem{LSTM.Peephole.Weights}
F.~Gers, N.~Schraudolph, and J.~Schmidhuber, ``Learning precise timing with
  lstm recurrent networks,'' \emph{Journal of Machine Learning Research},
  vol.~3, pp. 115--143, 2003.

\bibitem{CNN.Pooling.Lp}
P.~Sermanet, S.~Chintala, and Y.~LeCun, ``Convolutional neural networks applied
  to house numbers digit classification,'' in \emph{ICPR}, 2012, pp.
  3288--3291.

\bibitem{CNN.Pooling.Stochastic}
M.~Zeiler and R.~Fergus, ``Stochastic pooling for regularization of deep
  convolutional neural networks,'' in \emph{ICLR}, 2013.

\bibitem{NIN}
M.~Lin, Q.~Chen, and Y.~S., ``Network in network,'' in \emph{ICLR}, 2014.

\bibitem{LSTM.MultiDimensions}
A.~Graves, S.~Fern{\'{a}}ndez, and J.~Schmidhuber, ``Multi-dimensional
  recurrent neural networks,'' in \emph{International Conference on Artificial
  Neural Networks}, 2007.

\bibitem{LSTM.CNNInput}
S.~Sonderby, C.~Sonderby, H.~Nielsen, and O.~Winther, ``Convolutional lstm
  networks for subcellular localization of proteins,'' 2015, arXiv:1503.01919.

\bibitem{HKUST}
Y.~Liu, P.~Fung, Y.~Yang, C.~Cieri, S.~Huang, and D.~Graff, ``Hkust/mts: A very
  large scale mandarin telephone speech corpus,'' in \emph{ISCSLP}, 2006, pp.
  724--735.

\bibitem{LSTMRNNAM}
H.~Sak, A.~Senior, and F.~Beaufays, ``Long short-term memory recurrent neural
  network architectures for large scale acoustic modeling,'' in
  \emph{Interspeech}, 2014, pp. 338--342.

\bibitem{BPTT}
R.~Williams and J.~Peng, ``An efficient gradient-based algorithm for online
  training of recurrent neural network trajectories,'' \emph{Neural
  Computation}, vol.~2, pp. 490--501, 1990.

\bibitem{ASGD1}
R.~Orm{\'{a}}ndi, I.~Heged{\"{u}}s, and M.~Jelasity, ``Asynchronous
  peer-to-peer data mining with stochastic gradient descent,'' \emph{Lecture
  Notes in Computer Science}, pp. 528--540, 2011.

\bibitem{ASGD2}
S.~Zhang, C.~Zhang, Z.~You, R.~Zheng, and B.~Xu, ``Asynchronous stochastic
  gradient descent for dnn training,'' in \emph{ICASSP}, 2013, pp. 6660--6663.

\bibitem{RNN.difficulty}
R.~Pascanu and Y.~Bengio, ``On the difficulty of training recurrent neural
  networks,'' 2012, arXiv:1211.5063.

\bibitem{LSTM.Maxout}
X.~Li and X.~Wu, ``Improving long short-term memory networks using maxout units
  for large vocabulary speech recognition,'' in \emph{ICASSP}, 2015.

\bibitem{ReLU}
M.~Zeiler, M.~Ranzato, R.~Monga, M.~Mao, K.~Yang, Q.~Le, P.~Nguyen, A.~Senior,
  V.~Vanhouche, J.~Dean, and G.~Hinton, ``On rectified linear units for speech
  processing,'' in \emph{ICASSP}, 2013, pp. 3517--3521.

\bibitem{Relu2}
G.~Dahl, T.~Sainath, and G.~Hinton, ``Improving deep neural networks for lvcsr
  using rectified linear units and dropout,'' in \emph{ICASSP}, 2013, pp.
  8609--8613.

\bibitem{MaxoutAm1}
M.~Cai, Y.~Shi, and J.~Liu, ``Deep maxout neural networks for speech
  recognition,'' in \emph{ASRU}, 2013, pp. 291--296.

\bibitem{MaxoutAm2}
Y.~Miao, S.~Rawat, and F.~Metze, ``Deep maxout networks for low resource speech
  recognition,'' in \emph{ASRU}, 2013, pp. 398--403.

\end{thebibliography}


\end{document}